\documentclass[10pt,conference]{IEEEtran}
\IEEEoverridecommandlockouts
\usepackage{cite}
\usepackage{amsmath,amssymb,amsfonts}
\usepackage{algorithmic}
\usepackage{graphicx}
\usepackage{textcomp}
\usepackage{xcolor}

\usepackage{marvosym}

\usepackage{amssymb,amsfonts}
\usepackage{algorithmic}
\usepackage{textcomp}
\usepackage{xcolor}
\usepackage{booktabs, multicol, multirow, threeparttable}

\usepackage{tipa}
\usepackage{makecell}

\definecolor{Highlight}{rgb}{0.12,0.49,0.85}
\usepackage{caption}
\usepackage{marvosym}

\newcolumntype{?}{!{\vrule width 0.8pt}}

\usepackage[capitalize]{cleveref}
\crefname{section}{Sec.}{Secs.}
\Crefname{section}{Section}{Sections}
\Crefname{table}{Table}{Tables}
\crefname{table}{Tab.}{Tabs.}

\definecolor{bl}{rgb}{0.25, 0.5, 0.9}

\usepackage{xspace}
\newcommand{\NAME}{ReGuider\xspace}

\def\BibTeX{{\rm B\kern-.05em{\sc i\kern-.025em b}\kern-.08em
    T\kern-.1667em\lower.7ex\hbox{E}\kern-.125emX}}
\begin{document}

\title{Forecasting with Guidance: Representation-Level Supervision for Time Series Forecasting
}

\author{
\IEEEauthorblockN{Jiacheng Wang}
\IEEEauthorblockA{\textit{Xijing University}\\
2408540402040@stu.xijing.edu.cn \\
ORCID: 0009-0003-2474-4223}
\and
\IEEEauthorblockN{Liang Fan \quad \textnormal{and} \quad Baihua Li}
\IEEEauthorblockA{\textit{Loughborough University}\\
\{L.Fan, B.Li\}@lboro.ac.uk \\
ORCID: 0000-0003-1464-8353, 0000-0002-0277-3651}
\and
\IEEEauthorblockN{Luyan Zhang\textsuperscript{*}}
\IEEEauthorblockA{\textit{Northeastern University}\\
zhang.luya@northeastern.edu \\
ORCID: 0009-0008-7385-373X}
}

\maketitle

\begin{abstract}
Nowadays, time series forecasting is predominantly approached through the end-to-end training of deep learning architectures using error-based objectives. While this is effective at minimizing average loss, it encourages the encoder to discard informative yet extreme patterns. This results in smooth predictions and temporal representations that poorly capture salient dynamics. To address this issue, we propose \NAME, a plug-in method that can be seamlessly integrated into any forecasting architecture. \NAME leverages pretrained time series foundation models as semantic teachers. During training, the input sequence is processed together by the target forecasting model and the pretrained model. Rather than using the pretrained model's outputs directly, we extract its intermediate embeddings, which are rich in temporal and semantic information, and align them with the target model's encoder embeddings through representation-level supervision. This alignment process enables the encoder to learn more expressive temporal representations, thereby improving the accuracy of downstream forecasting. Extensive experimentation across diverse datasets and architectures demonstrates that our \NAME consistently improves forecasting performance, confirming its effectiveness and versatility.
\end{abstract}


\section{Introduction}
\label{sec:intro}

Time series forecasting (TSF) is central to many real-world applications, including finance~\cite{fintsb},  healthcare~\cite{gul2009health}, and climate science~\cite{weather_forecast}. The recent success of deep learning has brought substantial advances to the field, with architectures such as graph networks~\cite{timefilter,mtgnn}, Linear-based models~\cite{tide,dlinear}, and transformers~\cite{timebridge,fedformer} demonstrating strong predictive capabilities. By automatically extracting complex temporal dependencies, deep learning models have surpassed classical statistical approaches and become the predominant choice for modern forecasting tasks. However, achieving accurate and robust predictions across diverse domains remains a fundamental challenge.

Most deep learning approaches~\cite{itransformer,autoformer,patchtst} to time series forecasting rely solely on error-based objectives such as mean squared error (MSE) and mean absolute error (MAE). 
While these objectives optimize predictive accuracy directly, they provide the encoder with limited guidance on how to capture rich temporal dependencies. Consequently, models often reduce errors by averaging predictions, which can result in the neglect of outlier events and the formation of overly smoothed representations~\cite{amd}. 
This issue is particularly evident in the learned embeddings, which fail to encode sufficient temporal semantics. 
Such latent representations are often ``semantically impoverished'' as they capture the trend but lose the underlying generative dynamics of the system.

We argue that the key to improving forecasting performance lies not in designing increasingly complex architectures, but in incorporating external semantic supervision. Moreover, time series foundation models~\cite{chronos,timemoe,timer}, trained on large-scale and diverse data, learn temporal representations that more faithfully capture fine-grained and semantically rich patterns. To address the limitations of error-only supervision, we propose explicitly guiding the encoder using representations extracted from time-series foundation models, enabling it to learn more meaningful temporal abstractions. The core idea is to enrich the embeddings of forecasting methods through external semantic supervision, thereby enhancing their representational capacity without increasing architectural complexity.

Technically, we propose \NAME, a representation-level supervision plug-in designed to enhance time series forecasting. The central concept involves leveraging pretrained time series foundation models as semantic teachers. During training, both the target forecasting model and the pretrained model process the same input sequence. Instead of using the pretrained model’s prediction head, we extract its intermediate embeddings, which encode rich temporal dependencies and semantic structures. These embeddings are then aligned with the encoder representations of the target model, thereby encouraging the encoder to learn more expressive and temporally coherent representations. \NAME is model-agnostic, enabling it to be seamlessly integrated into a wide range of TSF methods without altering their original structure.

In summary, this work makes the following contributions:
\begin{itemize}
    \item We identify the limitation of error-only supervision in deep learning-based forecasting and propose to enhance temporal embeddings through external semantic guidance.
    \item We develop \NAME, a plug-in method that aligns encoder representations with pretrained time series foundation models, enriching the temporal semantics of learned embeddings.
    \item We conduct extensive experiments across diverse datasets and architectures, demonstrating that \NAME consistently improves forecasting accuracy and generalizes effectively to different backbone models.
\end{itemize}

\section{Background and Related Work}

\subsection{Deep Learning in Time Series Forecasting}
Deep learning has become the dominant paradigm in time series forecasting, with RNNs, CNNs, GNNs, and Transformers all demonstrating strong empirical results~\cite{autoformer, finmamba, timesnet, micn, revin}.  Recently, the community has started training large scale time series foundation models using hierarchical transformers or masked autoencoders that have been pre-trained on millions of sequences~\cite{timemoe, timer, calf, chronos}.  The immense capacity and extensive pre-training of these models enable them to capture universal temporal dynamics, ranging from short-term seasonality to long-term trends. Current practice involves either freezing or lightly fine-tuning these models for direct prediction or few-shot adaptation. However, we exploit their internal representations as repositories of high-quality temporal knowledge to enhance any downstream forecaster. 

\subsection{Representation Learning}

Representation learning is essential for enabling models to capture informative and transferable features. Across domains, it has been utilized to impose inductive biases that extend beyond simple task losses. For instance, in diffusion models, it is applied to make noise patterns more structured and controllable, improving generation quality and stability~\cite{repa}. In vision and language, aligning latent representations with pretrained models has proven effective in enriching feature spaces and boosting downstream performance~\cite{vavae}. 

Within time series forecasting, traditionally, TSF models used supervised encoders to extract features for point-wise prediction~\cite{autoformer,timemixer,timealign,etsformer,tide}. However, these are prone to ``representation collapse'' when driven solely by MSE, filtering out critical regime shifts to minimize average loss. While self-supervised learning (SSL) and contrastive paradigms attempt to mitigate this, they often rely on heuristic augmentations and remain limited by the scale of individual datasets.

The emergence of Time Series Foundation Models (TSFMs) has redefined this landscape. Pretrained on billions of sequences, TSFMs~\cite{chronos,timer,timexl,timesfm,chronos2} develop a ``universal temporal vocabulary'' that captures nuanced seasonality and structural dependencies. ReGuider bridges the gap between these high-capacity models and efficient task-specific predictors by using TSFM embeddings as a ``semantic gold standard'' for alignment.
This representation-level supervision can highlight long-term seasonality, abrupt regime shifts, and inter-variable relations that are often overlooked by error-driven objectives, thereby leading to more accurate and robust predictions.

\section{Method}

\subsection{Problem Statement}
The goal of TSF is to predict a future sequence $Y \in \mathbb{R}^{C \times T}$ with horizon $T$ from a past sequence $X \in \mathbb{R}^{C \times L}$ of length $L$, where $C$ denotes the number of variables.

\begin{figure}[!t]
    \centering
    \includegraphics[width=\linewidth]{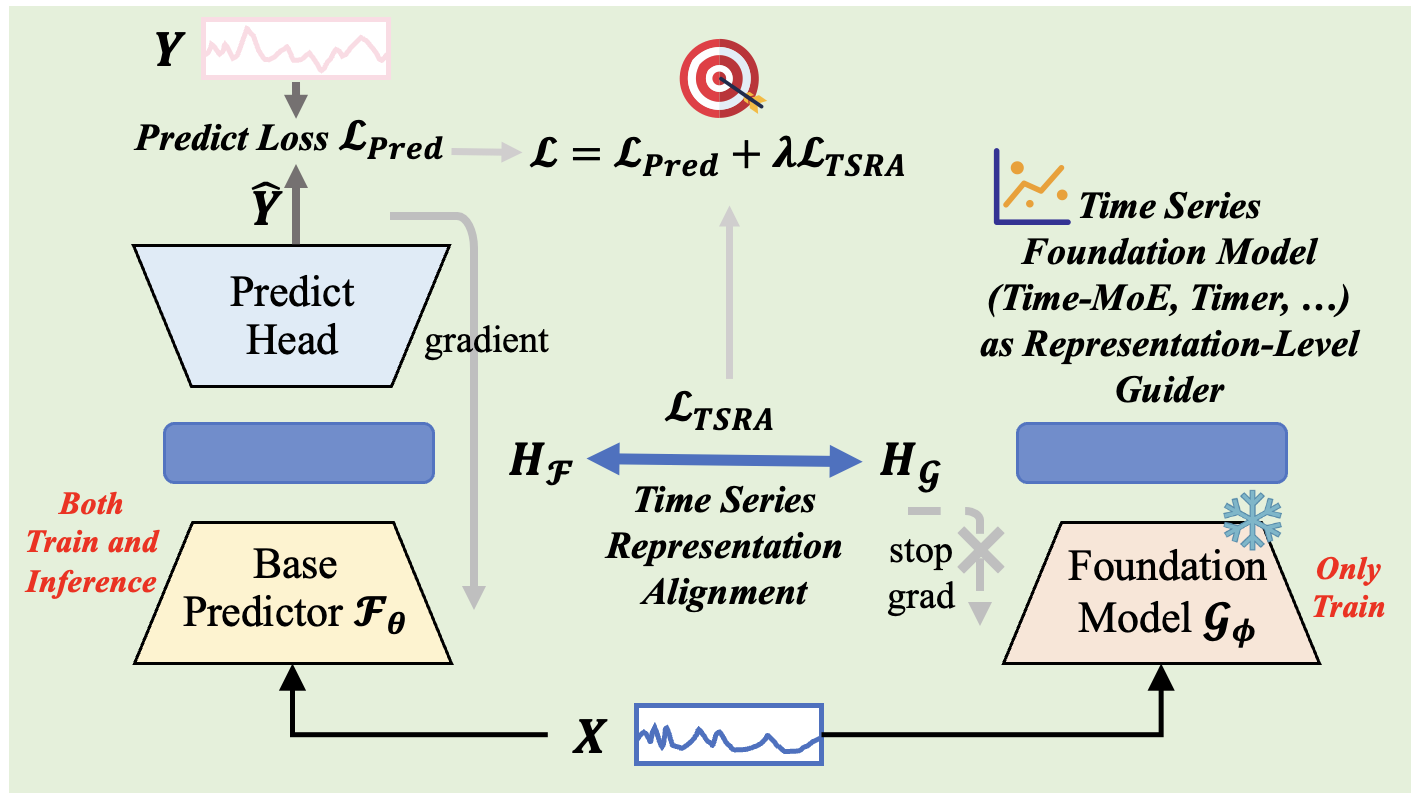}
    \caption{Overall architecture of \NAME, which consists of the base predictor $\mathcal{F}_{\theta}(\cdot)$ and the foundation model $\mathcal{G}_{\phi}(\cdot)$, serving as a representation guide.}
    \label{fig:pipeline}
\end{figure}

\subsection{Architecture of \NAME}
As shown in \cref{fig:pipeline}, \NAME is designed to improve time series forecasting by enriching encoder representations through supervision with pre-trained foundation models. 
Specifically, given an input sequence $X \in \mathbb{R}^{C \times L}$, it is processed through two parallel pathways: (1) the base predictor $\mathcal{F}_{\theta}(\cdot)$, representing the forecasting model to be trained, and (2) the foundation model $\mathcal{G}_{\phi}(\cdot)$, serving as a representation guide. The base predictor $\mathcal{F}_{\theta}(\cdot)$ encodes $X$ into a latent representation $H_f$ before passing it to the prediction head to generate an estimate of the target variable, denoted by $Y$. For the guider, rather than using the final prediction output of $\mathcal{G}_{\phi}(\cdot)$, we extract its intermediate embedding $H_g$ from the encoder. This captures the rich temporal patterns and high-level semantics learned during large-scale pretraining.
We then introduce a representation supervision objective to minimise the distance between $H_f$ and $H_g$. This guides the encoder of the base predictor to incorporate the temporal dependencies and semantic structures present in the pretrained $\mathcal{G}_{\phi}(\cdot)$. 

This supervision is seamlessly integrated into the training process alongside the standard forecasting loss, enabling the model to learn to minimize predictive error and produce embeddings that align with stronger temporal representation space simultaneously. This framework is also model-agnostic. \NAME does not alter the backbone framework or inference process, making it applicable to various TSF models.

\renewcommand{\arraystretch}{1.2}
\begin{table*}[!t]
\scriptsize
\centering
\setlength{\tabcolsep}{2pt}
\begin{threeparttable}
\resizebox{\textwidth}{!}{
\begin{tabular}{c|c|cccc?cccc?cccc?cccc}
\toprule

 \multicolumn{2}{c}{Models} & \multicolumn{2}{c}{iTransfomer~\cite{itransformer}} & \multicolumn{2}{c?}{+ \textbf{\NAME}} & \multicolumn{2}{c}{PatchTST~\cite{patchtst}} & \multicolumn{2}{c?}{+ \textbf{\NAME}} & \multicolumn{2}{c}{DLinear~\cite{dlinear}} & \multicolumn{2}{c?}{+ \textbf{\NAME}} & \multicolumn{2}{c}{TimeMixer~\cite{timemixer}} & \multicolumn{2}{c}{+ \textbf{\NAME}} \\
\cmidrule(lr){3-4} \cmidrule(lr){5-6} \cmidrule(lr){7-8} \cmidrule(lr){9-10}  \cmidrule(lr){11-12} \cmidrule(lr){13-14} \cmidrule(lr){15-16} \cmidrule(lr){17-18}
 \multicolumn{2}{c}{Metric} & MSE & \multicolumn{1}{c}{MAE} & MSE & MAE & MSE & \multicolumn{1}{c}{MAE} & MSE & MAE & MSE & \multicolumn{1}{c}{MAE} & MSE & MAE & MSE & \multicolumn{1}{c}{MAE} & MSE & MAE \\

\toprule

\multirow{4}{*}{\rotatebox[origin=c]{90}{ETTh1}} 

& 96  & 0.386 & 0.405 & \textbf{0.377} & \textbf{0.398} & 0.414 & 0.419 & \textbf{0.382} & \textbf{0.384} & 0.386 & 0.400 & \textbf{0.368} & \textbf{0.390} & 0.375 & 0.400 & \textbf{0.366} & \textbf{0.393} \\

& 192 & 0.441 & 0.436 & \textbf{0.427} & \textbf{0.426} & 0.460 & 0.445 & \textbf{0.424} & \textbf{0.425} & 0.437 & 0.432 & \textbf{0.402} & \textbf{0.413} & 0.429 & 0.421 & \textbf{0.422} & \textbf{0.419} \\

& 336 & 0.487 & 0.458 & \textbf{0.475} & \textbf{0.452} & 0.501 & 0.466 & \textbf{0.462} & \textbf{0.441} & 0.481 & 0.459 & \textbf{0.448} & \textbf{0.438} & 0.484 & 0.458 & \textbf{0.458} & \textbf{0.434} \\

& 720 & 0.503 & 0.491 & \textbf{0.486} & \textbf{0.480} & 0.500 & 0.488 & \textbf{0.478} & \textbf{0.471} & 0.519 & 0.516 & \textbf{0.487} & \textbf{0.485} & 0.498 & 0.482 & \textbf{0.475} & \textbf{0.465} \\

\midrule

\multirow{4}{*}{\rotatebox[origin=c]{90}{ETTh2}} 

& 96  & 0.297 & 0.349 & \textbf{0.289} & \textbf{0.343} & 0.302 & 0.348 & \textbf{0.293} & \textbf{0.338} & 0.333 & 0.387 & \textbf{0.320} & \textbf{0.361} & 0.289 & 0.341 & \textbf{0.382} & \textbf{0.334} \\

& 192 & 0.380 & 0.400 & \textbf{0.373} & \textbf{0.392} & 0.388 & 0.400 & \textbf{0.374} & \textbf{0.387} & 0.477 & 0.476 & \textbf{0.406} & \textbf{0.424} & 0.372 & 0.392 & \textbf{0.358} & \textbf{0.384} \\

& 336 & 0.428 & 0.432 & \textbf{0.415} & \textbf{0.427} & 0.426 & 0.433 & \textbf{0.412} & \textbf{0.421} & 0.594 & 0.541 & \textbf{0.453} & \textbf{0.455} & 0.386 & 0.414 & \textbf{0.379} & \textbf{0.410} \\

& 720 & 0.427 & 0.445 & \textbf{0.420} & \textbf{0.441} & 0.431 & 0.446 & \textbf{0.418} & \textbf{0.429} & 0.831 & 0.657 & \textbf{0.596} & \textbf{0.541} & 0.412 & 0.434 & \textbf{0.406} & \textbf{0.427} \\

\midrule

\multirow{4}{*}{\rotatebox[origin=c]{90}{ETTm1}} 

& 96  & 0.334 & 0.368 & \textbf{0.327} & \textbf{0.361} & 0.329 & 0.367 & \textbf{0.322} & \textbf{0.358} & 0.345 & 0.372 & \textbf{0.336} & \textbf{0.368} & 0.320 & 0.357 & \textbf{0.316} & \textbf{0.351} \\

& 192 & 0.377 & 0.391 & \textbf{0.372} & \textbf{0.386} & 0.367 & 0.385 & \textbf{0.357} & \textbf{0.378} & 0.380 & 0.389 & \textbf{0.369} & \textbf{0.372} & 0.361 & 0.381 & \textbf{0.355} & \textbf{0.377} \\

& 336 & 0.426 & 0.420 & \textbf{0.412} & \textbf{0.409} & 0.399 & 0.410 & \textbf{0.388} & \textbf{0.399} & 0.413 & 0.413 & \textbf{0.395} & \textbf{0.398} & 0.390 & 0.404 & \textbf{0.385} & \textbf{0.396} \\

& 720 & 0.491 & 0.459 & \textbf{0.476} & \textbf{0.442} & 0.454 & 0.439 & \textbf{0.445} & \textbf{0.430} & 0.474 & 0.453 & \textbf{0.461} & \textbf{0.442} & 0.454 & 0.441 & \textbf{0.444} & \textbf{0.431} \\

\midrule

\multirow{4}{*}{\rotatebox[origin=c]{90}{ETTm2}} 

& 96  & 0.180 & 0.264 & \textbf{0.175} & \textbf{0.258} & 0.175 & 0.259 & \textbf{0.168} & \textbf{0.248} & 0.193 & 0.292 & \textbf{0.173} & \textbf{0.269} & 0.175 & 0.258 & \textbf{0.170} & \textbf{0.252} \\

& 192 & 0.250 & 0.309 & \textbf{0.242} & \textbf{0.300} & 0.241 & 0.302 & \textbf{0.234} & \textbf{0.287} & 0.284 & 0.362 & \textbf{0.263} & \textbf{0.348} & 0.237 & 0.299 & \textbf{0.233} & \textbf{0.296} \\

& 336 & 0.311 & 0.348 & \textbf{0.303} & \textbf{0.339} & 0.305 & 0.343 & \textbf{0.301} & \textbf{0.335} & 0.369 & 0.427 & \textbf{0.344} & \textbf{0.401} & 0.298 & 0.340 & \textbf{0.291} & \textbf{0.330} \\

& 720 & 0.412 & 0.407 & \textbf{0.401} & \textbf{0.396} & 0.402 & 0.400 & \textbf{0.386} & \textbf{0.392} & 0.554 & 0.522 & \textbf{0.472} & \textbf{0.493} & 0.391 & 0.396 & \textbf{0.387} & \textbf{0.390} \\

\midrule

\multirow{4}{*}{\rotatebox[origin=c]{90}{Weather}} 

& 96  & 0.174 & 0.214 & \textbf{0.168} & \textbf{0.207} & 0.177 & 0.218 & \textbf{0.165} & \textbf{0.212} & 0.196 & 0.255 & \textbf{0.175} & \textbf{0.234} & 0.163 & 0.209 & \textbf{0.160} & \textbf{0.203} \\

& 192 & 0.221 & 0.254 & \textbf{0.216} & \textbf{0.248} & 0.225 & 0.259 & \textbf{0.208} & \textbf{0.244} & 0.237 & 0.296 & \textbf{0.212} & \textbf{0.258} & 0.208 & 0.250 & \textbf{0.205} & \textbf{0.246} \\

& 336 & 0.278 & 0.296 & \textbf{0.267} & \textbf{0.286} & 0.278 & 0.297 & \textbf{0.253} & \textbf{0.286} & 0.283 & 0.335 & \textbf{0.268} & \textbf{0.317} & 0.251 & 0.287 & \textbf{0.248} & \textbf{0.384} \\

& 720 & 0.358 & 0.347 & \textbf{0.346} & \textbf{0.338} & 0.354 & 0.348 & \textbf{0.342} & \textbf{0.340} & 0.345 & 0.381 & \textbf{0.324} & \textbf{0.372} & 0.339 & 0.341 & \textbf{0.336} & \textbf{0.337} \\

\midrule

\multirow{4}{*}{\rotatebox[origin=c]{90}{ECL}} 

& 96  & 0.148 & 0.240 & \textbf{0.143} & \textbf{0.236} & 0.181 & 0.270 & \textbf{0.163} & \textbf{0.256} & 0.197 & 0.282 & \textbf{0.166} & \textbf{0.269} & 0.153 & 0.247 & \textbf{0.152} & \textbf{0.245} \\

& 192 & 0.162 & 0.253 & \textbf{0.158} & \textbf{0.248} & 0.188 & 0.274 & \textbf{0.169} & \textbf{0.263} & 0.196 & 0.285 & \textbf{0.179} & \textbf{0.277} & 0.166 & \textbf{0.256} & \textbf{0.164} & \textbf{0.256} \\

& 336 & 0.178 & 0.269 & \textbf{0.173} & \textbf{0.265} & 0.204 & 0.293 & \textbf{0.195} & \textbf{0.286} & 0.209 & 0.301 & \textbf{0.196} & \textbf{0.289} & 0.185 & 0.277 & \textbf{0.182} & \textbf{0.272} \\

& 720 & 0.225 & 0.317 & \textbf{0.209} & \textbf{0.298} & 0.246 & 0.324 & \textbf{0.227} & \textbf{0.309} & 0.245 & 0.333 & \textbf{0.214} & \textbf{0.318} & 0.225 & 0.310 & \textbf{0.222} & \textbf{0.308} \\

\midrule

\multirow{4}{*}{\rotatebox[origin=c]{90}{Traffic}} 

& 96  & 0.395 & 0.268 & \textbf{0.379} & \textbf{0.262} & 0.462 & 0.295 & \textbf{0.405} & \textbf{0.272} & 0.650 & 0.396 & \textbf{0.521} & \textbf{0.293} & 0.462 & 0.285 & \textbf{0.397} & \textbf{0.270} \\

& 192 & 0.417 & 0.276 & \textbf{0.402} & \textbf{0.268} & 0.466 & 0.296 & \textbf{0.420} & \textbf{0.278} & 0.598 & 0.370 & \textbf{0.546} & \textbf{0.308} & 0.473 & 0.296 & \textbf{0.422} & \textbf{0.281} \\

& 336 & 0.433 & 0.283 & \textbf{0.431} & \textbf{0.276} & 0.482 & 0.304 & \textbf{0.434} & \textbf{0.292} & 0.605 & 0.373 & \textbf{0.552} & \textbf{0.329} & 0.498 & 0.296 & \textbf{0.441} & \textbf{0.290} \\

& 720 & 0.467 & 0.302 & \textbf{0.454} & \textbf{0.295} & 0.514 & 0.322 & \textbf{0.474} & \textbf{0.310} & 0.645 & 0.394 & \textbf{0.568} & \textbf{0.343} & 0.506 & 0.313 & \textbf{0.483} & \textbf{0.297} \\

\bottomrule
\end{tabular}
}

\caption{Long term forecasting results with varying predict lengths $T \in \{96, 192, 336, 720\}$. The historical input length $L$ is fixed at $96$ for fair comparison. The best results are highlighted in \textbf{bold}.}
\label{tab::res}
\end{threeparttable}
\end{table*}

\subsection{Representation Alignment with Supervised Representations}

To enable the base predictor to learn richer temporal dependencies, \NAME introduces an auxiliary representation supervision objective that aligns the encoder embedding of the base predictor with that of a pretrained foundation model.

Formally, given an input sequence $X \in \mathbb{R}^{C \times L}$, the base predictor $\mathcal{F}_{\theta}$ encodes it into a latent representation:
\begin{equation}H_f = \mathcal{F}_{\theta}^{\text{enc}}(X),
\end{equation}
while the foundation model $\mathcal{G}_{\phi}$ encodes the same sequence into:
\begin{equation}H_g = \mathcal{G}_{\phi}^{\text{enc}}(X),
\end{equation}
where $H_f, H_g \in \mathbb{R}^{d}$ denote the embeddings before the prediction head, and $\theta, \phi$ are the parameters of the base predictor and foundation model, respectively.

We define the representation supervision loss as:
\begin{equation}
\mathcal{L}_{\text{TSRA}}(\theta, \phi) = \mathbb{E}_{X \sim \mathcal{D}} \big[ \text{sim}(H_f, H_g) \big],
\end{equation}
where $\text{sim}(\cdot, \cdot)$ is a similarity or distance function. Several options are possible:

\textbf{Euclidean distance:} 
\begin{equation}
    \text{sim}_{\ell_2}(H_f, H_g) = || H_f - H_g ||_2^2.
\end{equation}

\textbf{Cosine similarity:} 
\begin{equation}\text{sim}_{\cos}(H_f, H_g) = 1 - \frac{H_f^\top H_g}{||H_f||_2 \, ||H_g||_2}.
\end{equation}

\textbf{KL divergence:} 
\begin{equation}\text{sim}_{\text{KL}}(H_f, H_g) = D_{\text{KL}} \big( \sigma(H_f) \,||\, \sigma(H_g) \big),
\end{equation}
where $\sigma(\cdot)$ denotes the softmax function.

The overall training objective combines the standard forecasting loss with the representation supervision loss:

\begin{equation}
\mathcal{L}_{\text{total}} = \mathcal{L}_{\text{Pred}}(Y, \hat{Y}) + \lambda \, \mathcal{L}_{\text{TSRA}}(\theta, \phi),
\end{equation}
where $\lambda$ is a trade-off hyperparameter. This joint objective ensures that the predictor minimizes forecasting error while simultaneously learning embeddings guided by the pretrained foundation model.

A critical design choice in ReGuider is the asymmetric gradient flow. The parameters $\phi$ of the foundation model are frozen to preserve the universal temporal vocabulary:
\begin{equation}\theta^*, \psi^* = \arg\min_{\theta, \psi} \mathcal{L}_{Pred}(Y, \hat{Y}) + \lambda \mathcal{L}_{TSRA}(\tilde{H}_f, \text{sp}(H_g)),
\end{equation}
where $\text{sp}(\cdot)$ denotes the stop gradient operation. This ensures that the foundation model acts as a stationary semantic anchor, preventing the representation drift that often occurs in traditional co-training paradigms.

\subsection{Discussions}

Although ReGuider involves an architecture like teacher-student, it differs from traditional knowledge distillation (KD). Whereas conventional KD focuses on output alignment by mimicking the teacher’s final predictions or logits, ReGuider emphasizes representation alignment. For time series, we argue that the 'teacher's' value lies not in its specific forecast values, but in its universal temporal vocabulary. By aligning intermediate latent spaces, we avoid inheriting the teacher’s potential predictive biases and instead focus on enriching the student's structural understanding.

\renewcommand{\arraystretch}{1.3}
\begin{table}[!t]
\scriptsize
\centering
\setlength{\tabcolsep}{1.5pt}
\begin{threeparttable}
\resizebox{\linewidth}{!}{
\begin{tabular}{c|c|cccccc?cccccc}
\toprule

 \multicolumn{2}{c}{\multirow{2}{*}{Models}} & \multicolumn{6}{c}{iTransformer \cite{itransformer}} & \multicolumn{6}{c}{PatchTST \cite{patchtst}} \\ 

\cmidrule(lr){3-8} \cmidrule(lr){9-14}

 \multicolumn{2}{c}{} & \multicolumn{2}{c}{ED} & \multicolumn{2}{c}{KLD} & \multicolumn{2}{c?}{Cos. Sim.} & \multicolumn{2}{c}{ED} & \multicolumn{2}{c}{KLD} & \multicolumn{2}{c}{Cos. Sim.}  \\
\cmidrule(lr){3-4} \cmidrule(lr){5-6} \cmidrule(lr){7-8} \cmidrule(lr){9-10} \cmidrule(lr){11-12} \cmidrule(lr){13-14} 
 \multicolumn{2}{c}{Metric} & MSE & \multicolumn{1}{c}{MAE} & MSE & MAE & MSE & MAE & MSE & \multicolumn{1}{c}{MAE} & MSE & MAE & MSE & MAE \\

\toprule

\multirow{4}{*}{\rotatebox[origin=c]{90}{ETTm1}} 
& 96  & 0.327 & \textbf{0.361} & 0.334 & 0.369 & \textbf{0.325} & 0.362 & \textbf{0.322} & \textbf{0.358} & 0.329 & 0.364 & 0.324 & 0.360 \\
& 192 & \textbf{0.372} & \textbf{0.386} & 0.379 & 0.392 & 0.373 & 0.388 & \textbf{0.357} & \textbf{0.378} & 0.363 & 0.385 & 0.359 & \textbf{0.378} \\
& 336 & \textbf{0.412} & \textbf{0.409} & 0.421 & 0.418 & 0.415 & 0.411 & \textbf{0.388} & \textbf{0.399} & 0.395 & 0.404 & 0.390 & 0.403 \\
& 720 & \textbf{0.476} & \textbf{0.442} & 0.482 & 0.451 & 0.479 & 0.445 & \textbf{0.445} & \textbf{0.430} & 0.452 & 0.435 & 0.447 & 0.431 \\
\midrule

\multirow{4}{*}{\rotatebox[origin=c]{90}{Weather}} 
& 96  & \textbf{0.168} & \textbf{0.207} & 0.173 & 0.213 & 0.169 & \textbf{0.207} & \textbf{0.165} & 0.212 & 0.171 & 0.214 & 0.166 & \textbf{0.210} \\
& 192 & \textbf{0.216} & 0.248 & 0.224 & 0.252 & 0.218 & \textbf{0.247} & \textbf{0.208} & \textbf{0.244} & 0.214 & 0.246 & 0.209 & 0.245 \\
& 336 & \textbf{0.267} & \textbf{0.286} & 0.276 & 0.292 & 0.270 & 0.287 & \textbf{0.253} & 0.286 & 0.259 & 0.288 & 0.254 & \textbf{0.285} \\
& 720 & \textbf{0.346} & \textbf{0.338} & 0.354 & 0.345 & 0.347 & 0.339 & 0.342 & 0.340 & 0.349 & 0.342 & \textbf{0.341} & \textbf{0.339} \\
\midrule

\multirow{4}{*}{\rotatebox[origin=c]{90}{ECL}} 
& 96  & \textbf{0.143} & \textbf{0.236} & 0.151 & 0.241 & 0.145 & 0.238 & \textbf{0.163} & \textbf{0.256} & 0.167 & 0.259 & 0.165 & 0.258 \\
& 192 & \textbf{0.158} & 0.248 & 0.166 & 0.253 & 0.160 & \textbf{0.247} & \textbf{0.169} & 0.263 & 0.175 & 0.266 & 0.170 & \textbf{0.262} \\
& 336 & \textbf{0.173} & \textbf{0.265} & 0.182 & 0.271 & 0.177 & 0.266 & \textbf{0.195} & \textbf{0.286} & 0.201 & 0.291 & 0.198 & 0.288 \\
& 720 & \textbf{0.209} & 0.298 & 0.217 & 0.305 & 0.210 & \textbf{0.297} & \textbf{0.227} & \textbf{0.309} & 0.234 & 0.314 & 0.228 & 0.310 \\
\midrule

\multirow{4}{*}{\rotatebox[origin=c]{90}{Traffic}} 
& 96  & \textbf{0.379} & \textbf{0.262} & 0.382 & 0.265 & 0.381 & 0.265 & \textbf{0.405} & \textbf{0.272} & 0.406 & 0.273 & 0.410 & 0.275 \\
& 192 & \textbf{0.402} & \textbf{0.268} & 0.404 & 0.269 & 0.406 & 0.271 & \textbf{0.420} & 0.278 & 0.421 & \textbf{0.277} & 0.424 & 0.280 \\
& 336 & \textbf{0.431} & \textbf{0.276} & 0.433 & 0.278 & 0.433 & 0.280 & \textbf{0.434} & 0.292 & 0.435 & \textbf{0.290} & 0.437 & 0.294 \\
& 720 & \textbf{0.454} & \textbf{0.295} & 0.456 & 0.296 & 0.460 & 0.299 & \textbf{0.474} & 0.310 & 0.475 & \textbf{0.309} & 0.478 & 0.312 \\

\bottomrule
\end{tabular}
}
\caption{Results of Euclidean distance (ED), KL divergence (KLD), and Cosine Similarity (Cos. Sim.) as optimization objectives for representation supervision.}
\label{tab::alignment}
\end{threeparttable}
\end{table}

\section{Experiments}

\subsection{Setups}

\textbf{Dataset}. We evaluate the proposed \NAME model on 7 commonly used time series benchmark datasets: ETTh1, ETTh2, ETTm1, ETTm2, Weather, Electricity, and Traffic~\cite{timesnet}. Specifically, the ETT series (ETTh1, ETTh2, ETTm1, ETTm2) records power load and oil temperature from electricity transformers at both hourly and 15-minute resolutions. The Weather dataset includes 21 meteorological indicators collected every 10 minutes, representing a typical low-dimensional physical sensing task. Electricity (ECL) tracks the hourly power consumption of 321 clients, serving as a mid-dimensional benchmark for demand forecasting. Finally, the Traffic dataset monitors hourly road occupancy rates from 862 sensors, providing a high-dimensional challenge for capturing complex spatial-temporal inter-dependencies.
Consistent with classic work~\cite{informer,autoformer}, we use Mean Squared Error (MSE) and Mean Absolute Error (MAE) as performance evaluation metrics.

\textbf{Base Predictor and Base Representation Guider}. For base predictor $\mathcal{F}(\cdot)$, it can be any mainstream deep learning-based time series forecasting model. We select four widely recognized models in LTSF literature: 
iTransformer~\cite{itransformer}, PatchTST~\cite{patchtst}, DLinear~\cite{dlinear}, and TimeMixer~\cite{timemixer}. We compare their direct forecasting performance with their use as base predictors in our proposed \NAME model. Furthermore, for foundation model $\mathcal{G}(\cdot)$ for Self-Supervised representation, we choose Time-MoE$_\text{base}$~\cite{timemoe}.

\renewcommand{\arraystretch}{1.3}
\begin{table}[!t]
\scriptsize
\centering
\setlength{\tabcolsep}{1.5pt}
\begin{threeparttable}
\resizebox{\linewidth}{!}{
\begin{tabular}{c|c|cccccc?cccccc}
\toprule

 \multicolumn{2}{c}{\multirow{2}{*}{Models}} & \multicolumn{6}{c}{iTransformer \cite{itransformer}} & \multicolumn{6}{c}{PatchTST \cite{patchtst}} \\ 

\cmidrule(lr){3-8} \cmidrule(lr){9-14}

 \multicolumn{2}{c}{} & \multicolumn{2}{c}{Base} & \multicolumn{2}{c}{Large} & \multicolumn{2}{c?}{Ultra} & \multicolumn{2}{c}{Base} & \multicolumn{2}{c}{Large} & \multicolumn{2}{c}{Ultra}  \\
\cmidrule(lr){3-4} \cmidrule(lr){5-6} \cmidrule(lr){7-8} \cmidrule(lr){9-10} \cmidrule(lr){11-12} \cmidrule(lr){13-14} 
 \multicolumn{2}{c}{Metric} & MSE & \multicolumn{1}{c}{MAE} & MSE & MAE & MSE & MAE & MSE & \multicolumn{1}{c}{MAE} & MSE & MAE & MSE & MAE \\

\toprule

\multirow{4}{*}{\rotatebox[origin=c]{90}{ETTm1}} 
& 96  & \textbf{0.325} & \textbf{0.361} & 0.327 & \textbf{0.361} & 0.334 & 0.368 & 0.324 & \textbf{0.357} & \textbf{0.322} & 0.358 & 0.336 & 0.370 \\
& 192 & \textbf{0.370} & 0.387 & 0.372 & \textbf{0.386} & 0.379 & 0.392 & \textbf{0.355} & \textbf{0.376} & 0.357 & 0.378 & 0.383 & 0.395 \\
& 336 & 0.414 & 0.412 & \textbf{0.412} & \textbf{0.409} & 0.420 & 0.416 & \textbf{0.386} & 0.400 & 0.388 & \textbf{0.399} & 0.423 & 0.419 \\
& 720 & \textbf{0.475} & 0.443 & 0.476 & \textbf{0.442} & 0.486 & 0.450 & 0.447 & 0.432 & \textbf{0.445} & \textbf{0.430} & 0.490 & 0.452 \\
\midrule

\multirow{4}{*}{\rotatebox[origin=c]{90}{Weather}} 
& 96  & 0.169 & 0.208 & \textbf{0.168} & \textbf{0.207} & 0.170 & 0.209 & 0.166 & 0.211 & \textbf{0.165} & 0.212 & 0.171 & \textbf{0.210} \\
& 192 & 0.217 & \textbf{0.247} & \textbf{0.216} & 0.248 & 0.219 & 0.249 & \textbf{0.208} & \textbf{0.243} & \textbf{0.208} & 0.244 & 0.210 & 0.246 \\
& 336 & \textbf{0.265} & \textbf{0.285} & 0.267 & 0.286 & 0.270 & 0.287 & 0.254 & 0.287 & \textbf{0.253} & \textbf{0.286} & 0.255 & 0.288 \\
& 720 & \textbf{0.345} & 0.340 & 0.346 & \textbf{0.338} & 0.348 & 0.339 & 0.343 & 0.341 & \textbf{0.342} & \textbf{0.340} & 0.344 & 0.342 \\
\midrule

\multirow{4}{*}{\rotatebox[origin=c]{90}{ECL}} 
& 96  & 0.148 & 0.240 & \textbf{0.143} & \textbf{0.236} & 0.152 & 0.243 & 0.167 & 0.259 & 0.163 & 0.256 & \textbf{0.160} & \textbf{0.255} \\
& 192 & 0.161 & 0.250 & \textbf{0.158} & \textbf{0.248} & 0.165 & 0.253 & 0.172 & 0.266 & 0.169 & \textbf{0.263} & \textbf{0.168} & \textbf{0.263} \\
& 336 & 0.178 & 0.270 & \textbf{0.173} & \textbf{0.265} & 0.181 & 0.272 & 0.196 & 0.287 & 0.195 & 0.286 & \textbf{0.192} & \textbf{0.284} \\
& 720 & 0.211 & 0.301 & 0.209 & \textbf{0.298} & \textbf{0.208} & 0.300 & 0.229 & 0.312 & \textbf{0.227} & \textbf{0.309} & 0.228 & 0.311 \\
\midrule

\multirow{4}{*}{\rotatebox[origin=c]{90}{Traffic}} 
& 96  & 0.391 & 0.270 & 0.379 & \textbf{0.262} & \textbf{0.376} & 0.263 & 0.412 & 0.276 & \textbf{0.405} & 0.272 & \textbf{0.405} & \textbf{0.270} \\
& 192 & 0.414 & 0.274 & \textbf{0.402} & \textbf{0.268} & 0.403 & \textbf{0.268} & 0.422 & 0.280 & \textbf{0.420} & \textbf{0.278} & \textbf{0.420} & \textbf{0.278} \\
& 336 & 0.439 & 0.280 & \textbf{0.431} & 0.276 & 0.433 & \textbf{0.275} & 0.436 & 0.291 & 0.434 & 0.292 & \textbf{0.432} & \textbf{0.289} \\
& 720 & 0.468 & 0.299 & \textbf{0.454} & \textbf{0.295} & \textbf{0.454} & \textbf{0.295} & 0.476 & 0.311 & \textbf{0.474} & 0.310 & 0.475 & \textbf{0.309} \\

\bottomrule
\end{tabular}
}
\caption{Results of Time-MoE foundation model—base, large, and ultra—as representation guiders.}
\label{tab::diff_guider}
\end{threeparttable}
\end{table}

\begin{figure*}[!t]
    \centering
    \includegraphics[width=0.95\textwidth]{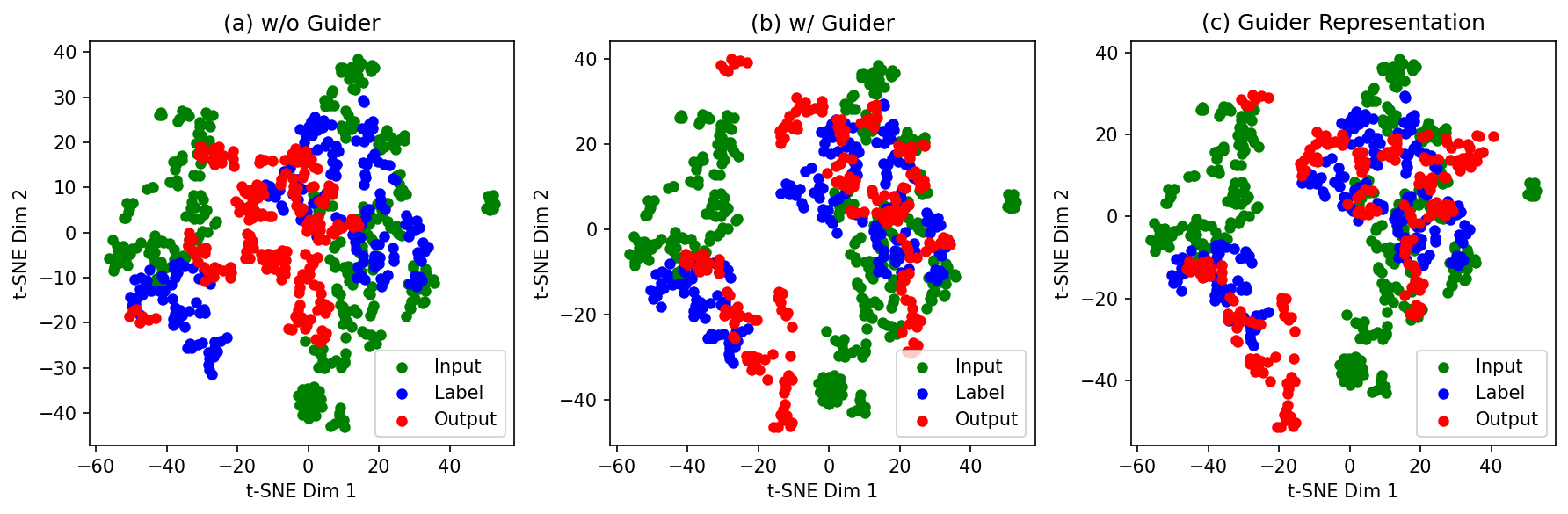}
    \caption{Visualisation of the iTransformer's and guider's representations.}
    \label{fig:vis}
\end{figure*}

\subsection{Main Results}

As shown in \cref{tab::res}, ReGuider consistently improves forecasting performance when applied to various backbone predictors, including Transformer- and Linear-based architectures. Across four representative backbones, our method improves forecasting accuracy by over $5\%$ on average, demonstrating its effectiveness in enriching temporal representations through representation-level supervision and confirming its generality as a seamless plug-in. This performance boost is particularly pronounced in high-dimensional datasets such as Traffic, which contains 862 variables. In these complex scenarios, the alignment with a foundation model's ``universal temporal vocabulary'' allows the base predictor to better capture intricate inter-variable couplings that are typically lost when training with error-only objectives, which tend to favor smoothed, uninformative averages.

Furthermore, ReGuider exhibits impressive stability as the forecasting horizon $T$ increases from 96 to 720. In short-term forecasting ($T=96$), the representation-level guidance assists the predictor in identifying sharp seasonality and abrupt regime shifts that point-wise losses often overlook. As the horizon extends to $T=720$, where standard models typically suffer from a widening drifting latent states, ReGuider acts as a semantic anchor. By enforcing alignment with the foundation model's stable embeddings ($H_g$), the student model sustains its predictive accuracy even at these challenging lengths, preventing the prediction from decaying toward a simple conditional mean.

\subsection{Model Analysis}

To further understand the behavior of \NAME, we investigate the following research questions:
\textbf{RQ1}: How should the distance between the base predictor’s embeddings and those of the foundation model be measured and optimized?
\textbf{RQ2}: How do different foundation models perform when serving as representation guiders?
\textbf{RQ3}: Does incorporating an additional guider significantly impact efficiency?
\textbf{RQ4}: Can we observe clear richer representation under guidance?

\textbf{RQ1. Distance metrics for supervision.}
We compare the use of Euclidean distance, KL divergence, and cosine similarity as optimization objectives for aligning the embeddings, and summarize the results in \cref{tab::alignment}. Of these three, Euclidean distance yields the best forecasting accuracy. This is because it directly measures point-wise closeness in the latent space, enforcing a tighter alignment between the two embedding distributions. In contrast, cosine similarity only constrains angular consistency without controlling magnitude, while KL divergence relies on distributional assumptions that may not hold in high-dimensional embedding spaces. Consequently, Euclidean distance provides the most stable and effective signal for representation supervision.

\textbf{RQ2. Effect of different foundation models.}
We also evaluate three versions of the Time-MoE foundation model — base, large, and ultra — as representation guiders. As shown in Table 1, results indicate that different pretrained representations offer complementary strengths. For instance, on relatively small datasets such as ETT, the base version provides competitive guidance, demonstrating that lightweight models can effectively transfer temporal semantics. However, on larger, more complex datasets such as Traffic, the ultra variant with the highest parameter count achieves the best performance, highlighting the benefit of scaling foundation models to capture broader temporal patterns. These results suggest that the guider chosen should be adapted to the scale and complexity of the target dataset.

\textbf{RQ3. Efficiency considerations.}
In terms of computational cost, \NAME only requires the foundation model to be invoked during training in order to extract intermediate embeddings. While this introduces a marginal increase in training time, it does not affect inference since the foundation model is no longer needed once alignment is complete. Consequently, the method incurs negligible overhead at deployment, ensuring that the benefits of representation-level supervision are realised without any additional inference cost.

\textbf{RQ4. Representation Visualization.}
To verify the effect of supervision, we feed a randomly selected window from the ETTm1 test set into both the vanilla iTransformer~\cite{itransformer} encoder and the \NAME-trained version of the same encoder.  We then reduced the dimensions using t-SNE.
As shown in \cref{fig:vis}, vanilla embeddings form a diffuse cloud with substantial overlap among trend classes, indicating weak temporal discrimination.  After \NAME supervision, the same encoder produces compact, well-separated clusters. This visual separation confirms that the guidance objective has transferred the foundation model’s rich temporal structure to the encoder, providing a more informative latent space.

\section{Conclusion}
In this study, we present \NAME, a representation-level supervision method in the form of a plug-in for TSF. \NAME enriches encoder embeddings by aligning them with representations extracted from pre-trained time series foundation models. This design allows forecasting architectures to capture richer temporal dependencies and semantic structures, delivering consistent performance enhancements across various backbones and datasets. Extensive experimentation confirms that \NAME is effective and efficient, enhancing accuracy without incurring additional inference costs. Ultimately, We believe that this framework underscores the potential of foundation models as universal representation guides, opening new avenues for semantically aware temporal modeling.

\section{LLM usage description}
Large Language Models (LLMs) were used solely as additional tools for refining the language. The authors had full autonomy over all aspects of the research, including its conceptualisation, experimental execution and data interpretation. No AI was involved in the core scientific processes or the derivation of conclusions.

\bibliographystyle{IEEEbib}
\bibliography{ref}

\end{document}